\documentclass[conference]{IEEEtran}
\IEEEoverridecommandlockouts
\usepackage{cite}
\usepackage{amsmath,amssymb,amsfonts}
\usepackage{algorithmic}
\usepackage{graphicx}
\usepackage{textcomp}
\usepackage{xcolor}
\usepackage{multirow}
\def\BibTeX{{\rm B\kern-.05em{\sc i\kern-.025em b}\kern-.08em
    T\kern-.1667em\lower.7ex\hbox{E}\kern-.125emX}}
\usepackage{subcaption}
\begin{document}

\title{A Deep Reinforcement Learning Blind AI in DareFightingICE\\
}

\makeatletter
\newcommand{\linebreakand}{%
  \end{@IEEEauthorhalign}
  \hfill\mbox{}\par
  \mbox{}\hfill\begin{@IEEEauthorhalign}
}
\makeatother

\author{\IEEEauthorblockN{Thai Van Nguyen, Xincheng Dai, Ibrahim Khan}
\IEEEauthorblockA{\textit{Graduate School of Information Science and Engineering} \\
\textit{Ritsumeikan University}\\
Kusatsu, Shiga, Japan \\
\{gr0557fv, gr0502pv, gr0556vx\}@ed.ritsumei.ac.jp}
\and
\IEEEauthorblockN{Ruck Thawonmas}
\IEEEauthorblockA{\textit{College of Information Science and Engineering} \\
\textit{Ritsumeikan University}\\
Kusatsu, Shiga, Japan \\
ruck@is.ritsumei.ac.jp}
\linebreakand 
\IEEEauthorblockN{Hai V. Pham}
\IEEEauthorblockA{\textit{School of Information and Communication Technology} \\
\textit{Hanoi University of Science And Technology}\\
Hanoi, Vietnam\\
haipv@soict.hust.edu.vn}
}

\maketitle

\begin{abstract}
This paper presents a deep reinforcement learning agent (AI) that uses sound as the input on the DareFightingICE platform at the DareFightingICE Competition in IEEE CoG 2022. In this work, an AI that only uses sound as the input is called blind AI. While state-of-the-art AIs rely mostly on visual or structured observations provided by their environments, learning to play games from only sound is still new and thus challenging. We propose different approaches to process audio data and use the Proximal Policy Optimization algorithm for our blind AI. We also propose to use our blind AI in evaluation of sound designs submitted to the competition and define two metrics for this task. The experimental results show the effectiveness of not only our blind AI but also the proposed two metrics. 
\end{abstract}


\section{Introduction}
Sounds in video games have been an important factor for a long time \cite{b1,b2,b3}. There are different types of sound effects in a game, ranging from UI sounds, ambient sounds, background music, and so on. These different types of sound effects improve the immersion of the human players. Human players use this sound input from the game for various tasks, such as, finding the location of an item or an enemy, and recognizing objects by their specific sounds. Sounds in a video game can help human players in different ways, but the question that we address in this paper is: Can AIs learn from sound in a video game? \par

Recent research has shown that an AI can use sound as an input to detect the location and direction of an object \cite{b4}. An AI that uses sound as an input together with other information is shown to perform better than those who do not \cite{b5}. However, research in AIs that use sound as the input is still in its premature stage, and previous research either focused on learning to play simplified games from only audio cues or used visual and other inputs like game data along with audio data in their AI. Our research introduces an AI that only uses sound as the input to play a fighting game called “DareFightingICE”. \par

DareFightingICE \cite{b6} is an enhanced version of FightingICE \cite{b7}, a fighting game, which has been used as the platform for the Fighting Game AI Competition series since 2013. DareFightingICE is an official competition in the 2022 IEEE Conference on Games. DareFightingICE has a “Sound-Only” option in which AIs only receive audio data as the input. \par

The contributions of this work are as follows: first, the creation of the first fighting game AI, an official sample AI in the competition, that uses only sound as the input (blind AI); second, our usage of the AI in evaluating the effectiveness of a given sound design regarding its in-game-event representation ability; and third, our work opens a new door to research into blind AIs.  \par

\section{Related Work}

\subsection{Artificial Intelligence techniques for sound}

Audio signal processing is one of the most important areas of Artificial Intelligence. In recent years, because deep learning has become more and more ubiquitous, it has been applied to audio processing and therefore has gained successes in applications such as speech recognition\cite{b8},\cite{b9} and text-to-speech\cite{b10}, \cite{b11}. One common way to apply audio processing in deep learning is to convert the audio data into images and then process them as with other images. This can be done by generating spectrograms, which are 2D images representing sequences of spectra with time and frequency along two axes and with color representing the strength of a frequency component. A spectrogram can be obtained by applying Short-Time Fourier Transform (STFT) to the audio signal, and the STFT of a signal can be calculated using Fast Fourier Transform (FFT). A Mel-frequency scale can also be applied to spectrograms to create Mel-spectrogram, which is better for human perception and is used in \cite{b8} and \cite{b9}. In our work, as audio encoders for our blind AI, we compare three types of transformations that use a convolutional neural network (CNN) containing two 1D convolutional layers, a combination of FFT and a 2-layer fully connected network (FCN), and a combination of Mel-spectrogram and a CNN containing two 2D convolutional layers, respectively.

\subsection{Game AI based on sound}
There have been a number of existing studies that focused on game playing AI using sound. Gaina and Stephenson\cite{b4} expanded the General Video Game AI framework to support sound and trained AIs to play the game using only sound as the input. Hegde et al.\cite{b5} extended the standard VizDoom framework \cite{b12} to provide the in-game sound to AIs and trained AIs in a series of increasingly complex scenarios to test the perception of sound. Results from these studies show the potential of research on an AI that learns to play games from sound. However, in these studies, AIs were trained with only one sound design and were not used to evaluate the effectiveness of sound designs in games. Therefore, our work, to the best of our knowledge, is the first time an AI is used to train with multiple sound designs and to evaluate their effectiveness.
\subsection{Proximal Policy Optimization}
The Proximal Policy Optimization (PPO) \cite{b13} algorithm has been a popular reinforcement learning approach in recent years. PPO provides a reliable Trust Region Policy Optimization based on previous policy gradient methods and outperforms traditional methods such as Q-learning. Its policy loss function is as follows:
\begin{equation}
    {L}_t^{CLIP}\theta = \hat{E}_t[min(\rho_t(\theta)\hat{A}_t, clip(\rho_t(\theta), 1-\epsilon,1+\epsilon)\hat{A}_t)]
\end{equation}
\begin{equation}
    \rho_t(\theta) = \frac{\pi_{\theta}(a_t|s_t)}{\pi_{\theta_{old}}(a_t|s_t)}
\end{equation}
\begin{equation}
    \hat{A}_t = \sigma_t + (\gamma\lambda)\sigma_{t+1} + ... + {(\gamma\lambda)}^{T-t+1}\sigma_{T-1}
\end{equation}
\begin{equation}
    \sigma_t = r_t + {\gamma}V_{\theta}(s_{t+1}) - V_{\theta}(s_t)
\end{equation}
In the equations above, $s_t$ and $a_t$ are the state and action at timestep $t$, respectively, $\pi_{\theta}(a_t|s_t)$ and $\pi_{\theta_{old}}(a_t|s_t)$ are the probability of $a_t$ given $s_t$ of the current policy and the previous policy, respectively. $V_{\theta}(s_t)$ is the value function of state $s_t$, and $\epsilon$, $\lambda$, $\gamma$ are clipping, Generalized Advantage Estimate (GAE) and discount factor, respectively. In the Fighting Game AI competition, PPO had been used in a number of studies \cite{b14,b15,b16} and achieved a remarkable success, especially as can be seen from the 2021 champion and runner-up both using PPO as part of their AI\footnote{https://www.ice.ci.ritsumei.ac.jp/~ftgaic/index-R.html}. In addition, PPO is outstanding in audio processing tasks, such as audio-based navigation in a multi-speaker environment \cite{b17} and semantic audio-visual navigation, where objects' sounds are consistent with their semantic meaning \cite{b18}. In this work, we use PPO to train our blind AI.

\subsection {Fighting Game AI}
Many state-of-the-art algorithms tested their performance in the Fighting Game AI Competition, including deep reinforcement learning (DRL). Kim et al. \cite{b14} created a fighting game AI agent using deep reinforcement learning with self-play and Monte Carlo Tree Search (MCTS), and they later proposed a reinforcement learning agent to tune the environment slightly by reusing latent space obtained from a different environment\cite{b16}. In 2020, Tang et al. \cite{b19} proposed a  method that combined the Rolling Horizon Evolution Algorithm with an opponent model and won the competition that year. In 2021, Liang et al. \cite{b15} extended the said work by proposing PPO with an Elo-based selection mechanism in which strong historical AIs are more frequently chosen during training. However, all these AIs use frame data provided by the competition until 2021. From 2022, the DareFightingICE competition \cite{b6}, using the titular game platform, has been launched where entry AIs are required to play the game from audio data only and have no access to frame data information. Therefore, our work is the first effort to train an AI to play a fighting game using only sound as the input.
\section{Methodology}
\subsection{Preprocessing}

Raw audio data provided by DareFightingICE are in the form of a vector ${\textit{\textbf{s}}} \in [-1,1]^n$ containing $n$ normalized audio samples. The original raw audio data size is 800 for each of the two channels (left and right), but each channel is padded with zeros so that it has 1024 samples for the sake of FFT in Java\cite{b6}. In our work, however, we choose to use only the 800 original samples for each channel. Motivated by the work of Hedge et al.\cite{b5}, we propose and compare three encoders to process the audio data, before feeding them to a deep neural network, in the following. Figure ~\ref{figEncoder} shows the architectures of these encoders, where all the parameters are based on previous work ~\cite{b5}, but empirically adapted for our work.
\subsubsection{1D-CNN}
The audio data ${\textit{\textbf{s}}}$ are downsampled by taking every $8^{th}$ sample and fed to two 1D convolution layers. Downsampling helps reduce the computational complexity. In the end, an audio-feature vector of 32x5 is obtained.
\subsubsection{FFT}
The input audio data ${\bf s}$ are transformed to frequency domain using FFT, and the FFT data are converted to the natural logarithm of the magnitudes $s{FFT} = logFFT({\textit{\textbf{s}}})\in R^{n/2}$. The resulting data are then downsampled and fed to a two-layer FCN. In the end, a one-dimensional audio-feature vector of 256 is obtained.
\subsubsection{Mel-spectrogram}
The input data ${\textit{\textbf{s}}}$ are transformed to frequency domain spectrogram with short-term Fourier transform (STFT), which is a sequence of Fourier transform of a windowed signal moved with a given hop. The frequencies are then processed with Mel scale. For hyperparameter setting, we choose a hop size of 10 ms, a window size of 25ms and 80 mel-frequency components. The spectrogram data are then fed to a network of two 2D convolutional layers. In the end, an audio-feature vector of 32x40x1 is obtained.
 \begin{figure*}
\centerline{\includegraphics[width=\textwidth]{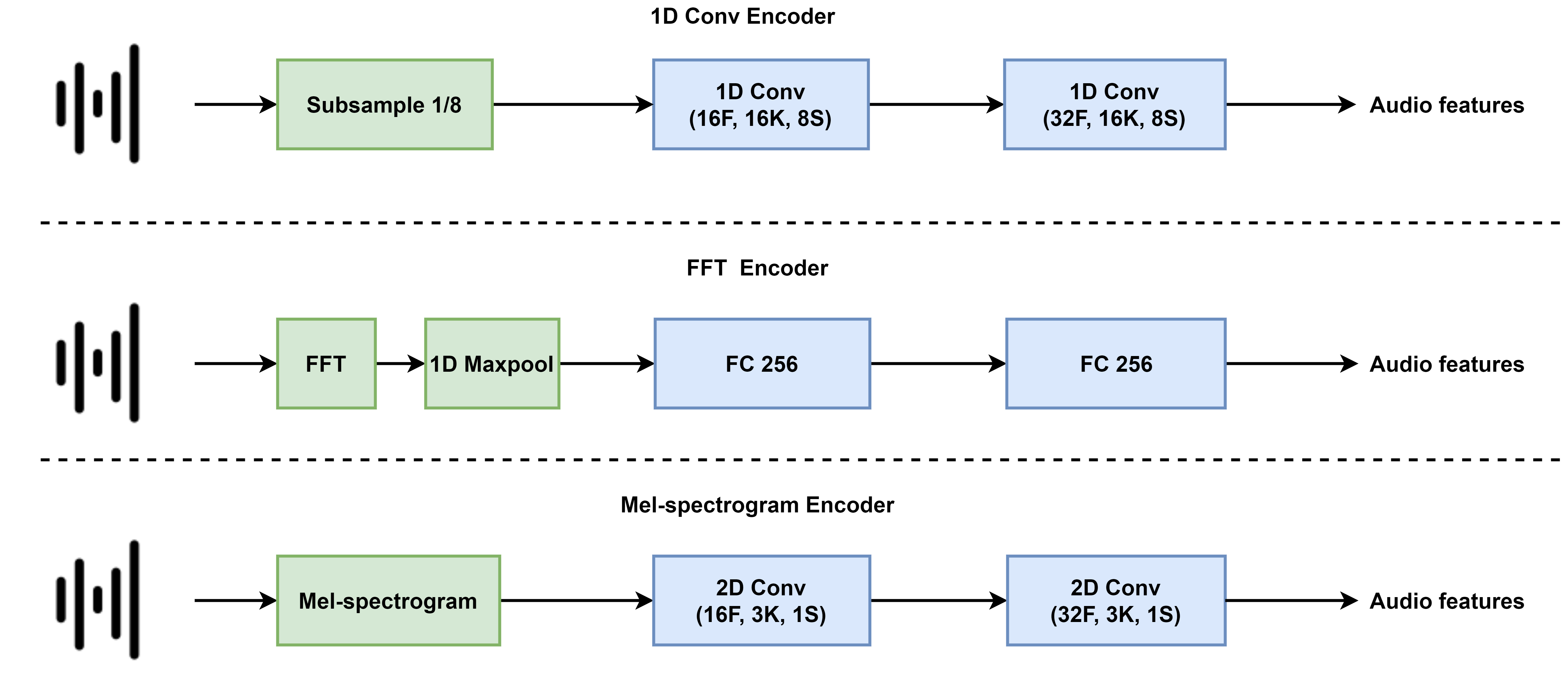}}
\caption{Audio encoders: 1D-CNN (top), FFT (middle), and Mel-spectrogram (bottom).}
\label{figEncoder}
\end{figure*}
\subsection{AI Design}
DRL refers to a growing set of powerful algorithms which use deep neural networks to learn in environments that have high dimensional states and actions. In our work, as stated earlier at the end of II-C, we use PPO whose architecture and reward are described in the following.
\subsubsection{Network Architecture}
Our model consists of a chosen audio encoder given in the previous section, a gated recurrent unit\cite{b20}, and a fully connected three-layered network to produce action probabilities. The fully connected layer network consists of three layers. The input of the network is the output of the audio encoder. There are two hidden layers, and each layer has 256 nodes. The output layer contains 40 nodes representing 40 actions\footnote{The actions in use consist of 2 throw-in-ground, 12 attack-in-ground, 3 skill-in-ground, 7 movement-in-ground, 2 guard-in-ground, 12 attack-in-air, and 2 skill-in-air actions.} reused from \cite{b15}, where the CROUCH action is omitted. We follow the PPO hyperparameters setting from \cite{b15}, as shown in Table~\ref{tblPPOParams}. The architecture of our AI is depicted in Fig.~\ref{figArchitecture}.
 \begin{figure*}
\centerline{\includegraphics[width=\textwidth]{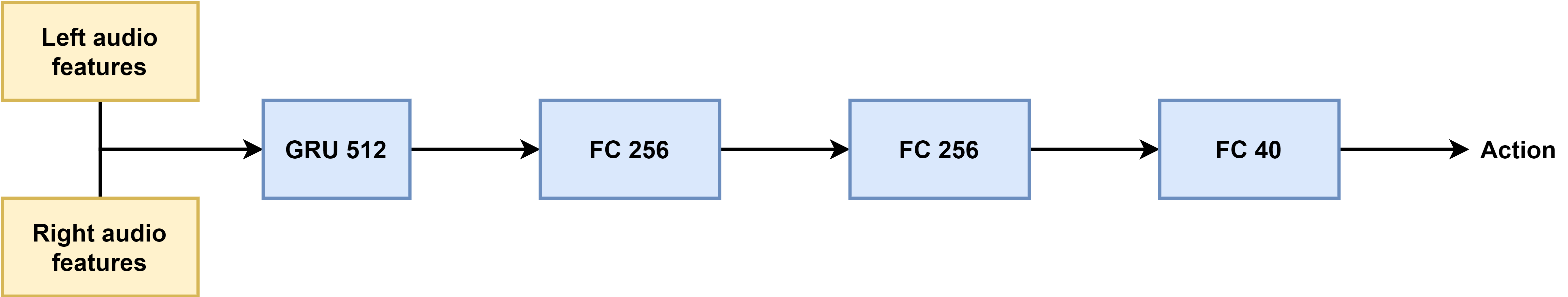}}
\caption{Our blind AI architecture.}
\label{figArchitecture}
\end{figure*}
\begin{table}[t!]
\caption{Hyperparameters setting of PPO.}
\label{tblPPOParams}
\begin{center}
\begin{tabular}{|c|c|}
\hline
Hyperparameter & Value\\
\hline
Adam step size & $3*10^{-4}$\\
\hline
Number of epochs for optimizing surrogate & 10 \\
\hline
Mini batch size & 64 \\
\hline
Discount (${\gamma}$) & 0.99 \\
\hline
GAE parameter (${\lambda}$) & 0.95 \\
\hline
GRU hidden units & 512 \\
\hline
\end{tabular}
\end{center}
\end{table}
\subsubsection{Reward Definition}
Following the recipe in previous work \cite{b21}, we define the reward function as follows:

\begin{equation}
    Reward = {Reward}_{t}^{offense} + {Reward}_{t}^{defense}
    \label{rewardStandard}
\end{equation}
\begin{equation}
    {Reward}_{t}^{offense} = {HP}_{t}^{opp} - {HP}_{t+1}^{opp}
    \label{rewardOffense}
\end{equation}
\begin{equation}
    {Reward}_{t}^{defense} = {HP}_{t+1}^{self} - {HP}_{t}^{self}
    \label{rewardDefense}
\end{equation}
where $t$ and $t+1$ represent the current frame step and the subsequent step, respectively; HP indicates the hit points of a character of interest and will be decreased if the character ($self$) receives damage from its opponent ($opp$), and in this work as well as the competition is initialized with a value of 400 at the beginning of a game round.
\subsection{Competition Metrics}
Here we propose two metrics for evaluating the effectiveness of a given sound design and/or a given audio encoder: $win_{ratio}$, and $avg HP_{diff}$. They are defined in a way that the more effective a sound design or an audio encoder is, the higher values these two metrics will be.
We first train three blind AIs, each using a different audio encoder. The opponent AI in use is a weakened version of the MCTS sample AI in the preceding competition, MctsAi65 discussed in Khan et al. \cite{b6}. Each training lasts 900 game rounds. MctsAi65 is an AI whose execution time is reduced to 6.5 ms and uses frame data as the input. Because MctsAi65 was selected in \cite{b6} for human evaluation, as it is not too strong for visually impaired players to play, we choose it as the opponent AI.

We then evaluate the fighting performance of each trained blind AI by making it fight against the aforementioned opponent AI for 90 rounds. The ratio of the number of wins\footnote{In the game, the round winner is either the one with a non-zero HP while its opponent's HP has reached zero or the one with the higher HP when the round-length limit of 60 s has reached.} over 90 rounds, Eqn. \eqref{equWinRate}, and the average HP difference at the end of a round between the trained AI and its opponent, Eqn. \eqref{HPDifference}, are then calculated.
\begin{equation}
    win_{ratio} = \frac {\textit{winning rounds}}{\textit{total rounds}}\label{equWinRate}
\end{equation}
\begin{equation}
    avg HP_{diff} = \frac {\textit{sum of }HP_r^{self} - HP_r^{opp}\textit{ for all r}}{\textit{total rounds}} \label{HPDifference}
\end{equation}
\section{Results and Discussions}

\begin{table}[t!]
\caption{Performance of our blind AI with different sound designs and different audio encoders and when fighting against the opponent AI.}
\label{tblFightResults}
\begin{center}
\begin{tabular}{|c|c|c|c|}
\hline
Sound design &Encoder & $win_{ratio}$ & $avg HP_{diff}$\\
\hline
DareFightingICE &1D-CNN & 0.33 & -28.83\\
\hline
DareFightingICE &FFT & 0.37 & -40.5\\
\hline
DareFightingICE &Mel-spectrogram & {\bf 0.63} & {\bf 37.07}\\
\hline
FightingICE v4.5 &1D-CNN & 0.5 & 4.5\\
\hline
FightingICE v4.5 &FFT & 0.51 & 3.25\\
\hline
FightingICE v4.5 &Mel-spectrogram & 0.57 & 21.94\\
\hline
\end{tabular}
\end{center}
\end{table}

\setcounter{figure}{3}
\renewcommand{\thefigure}{\arabic{figure}}
\begin{figure*}[t!]
    \centering
    \includegraphics[width=.5\linewidth]{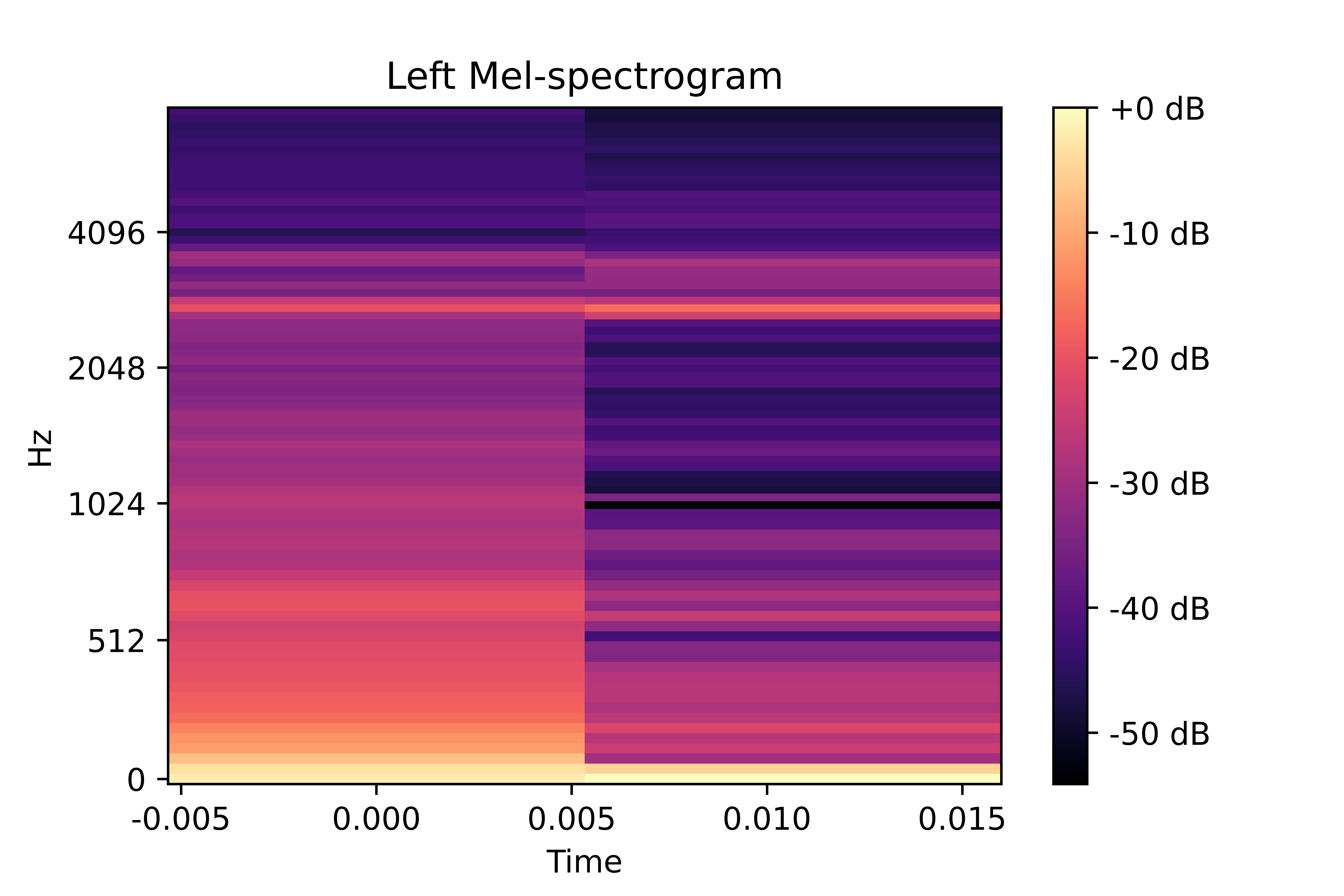}\hfill
    \includegraphics[width=.5\linewidth]{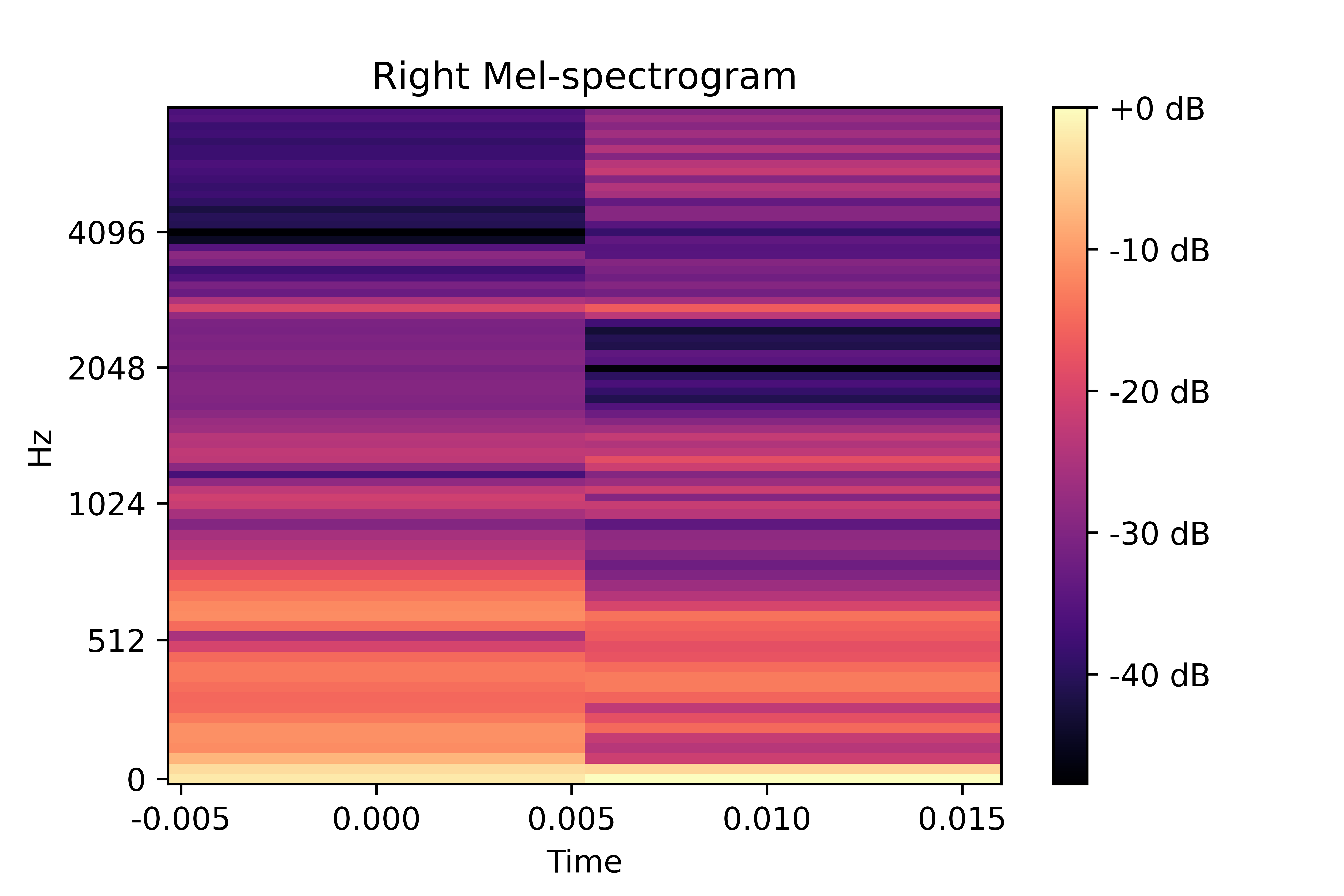}\hfill
    \caption{Mel-spectrogram results of audio data when the opponent AI attacks with a fireball in the DareFightingICE sound design.}
    \label{fig:fireballFFTDareFightingICE}
\end{figure*}
\begin{figure*}[t!]
    \centering
    \includegraphics[width=.5\linewidth]{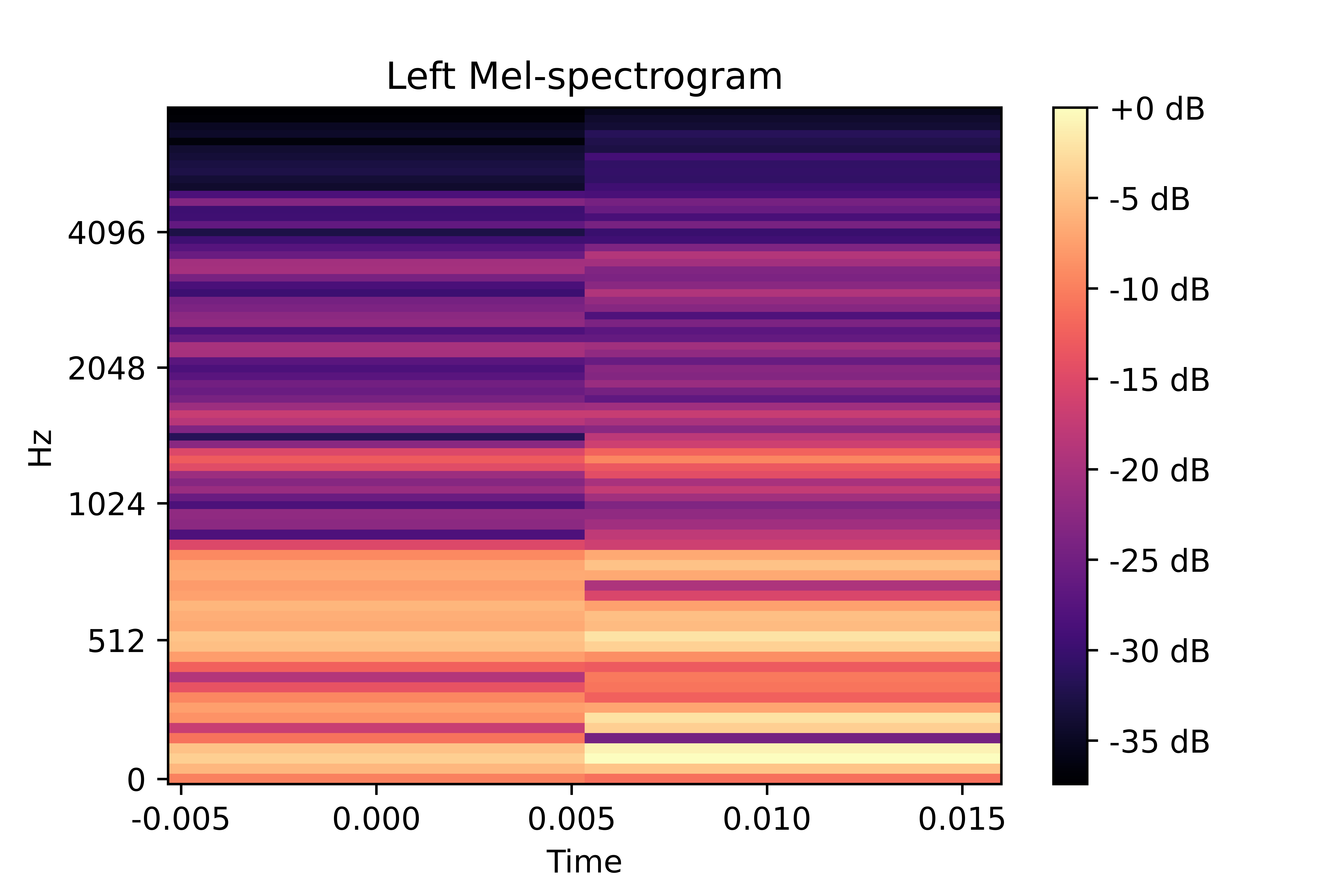}\hfill
    \includegraphics[width=.5\linewidth]{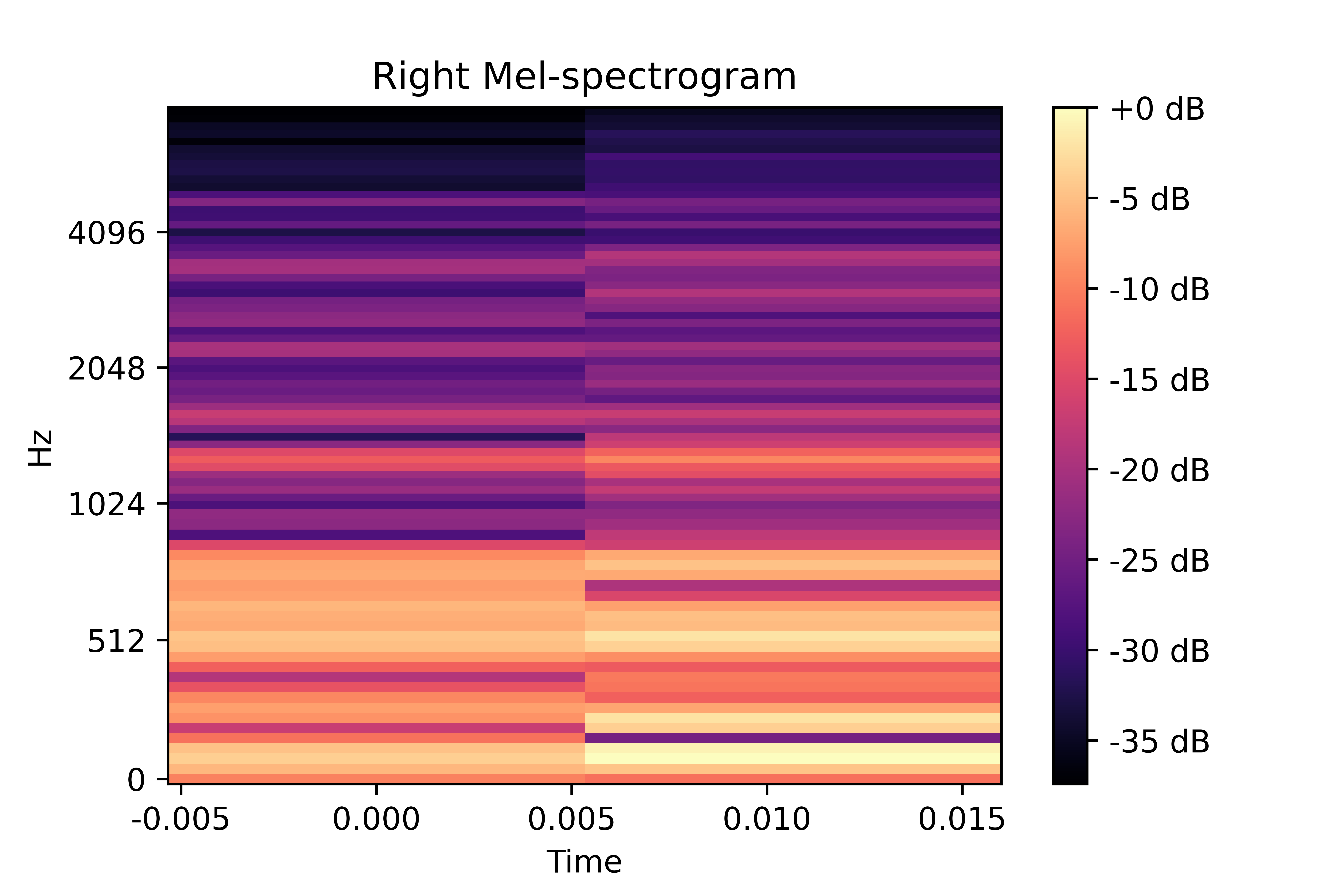}\hfill
    \caption{Mel-spectrogram results of audio data when the opponent AI attacks with a fireball in the FightingICE sound design.}
    \label{fig:fireballFFTFightingICE4.5}
\end{figure*}
\begin{figure*}[t!]
    \centering
    \includegraphics[width=.33\linewidth]{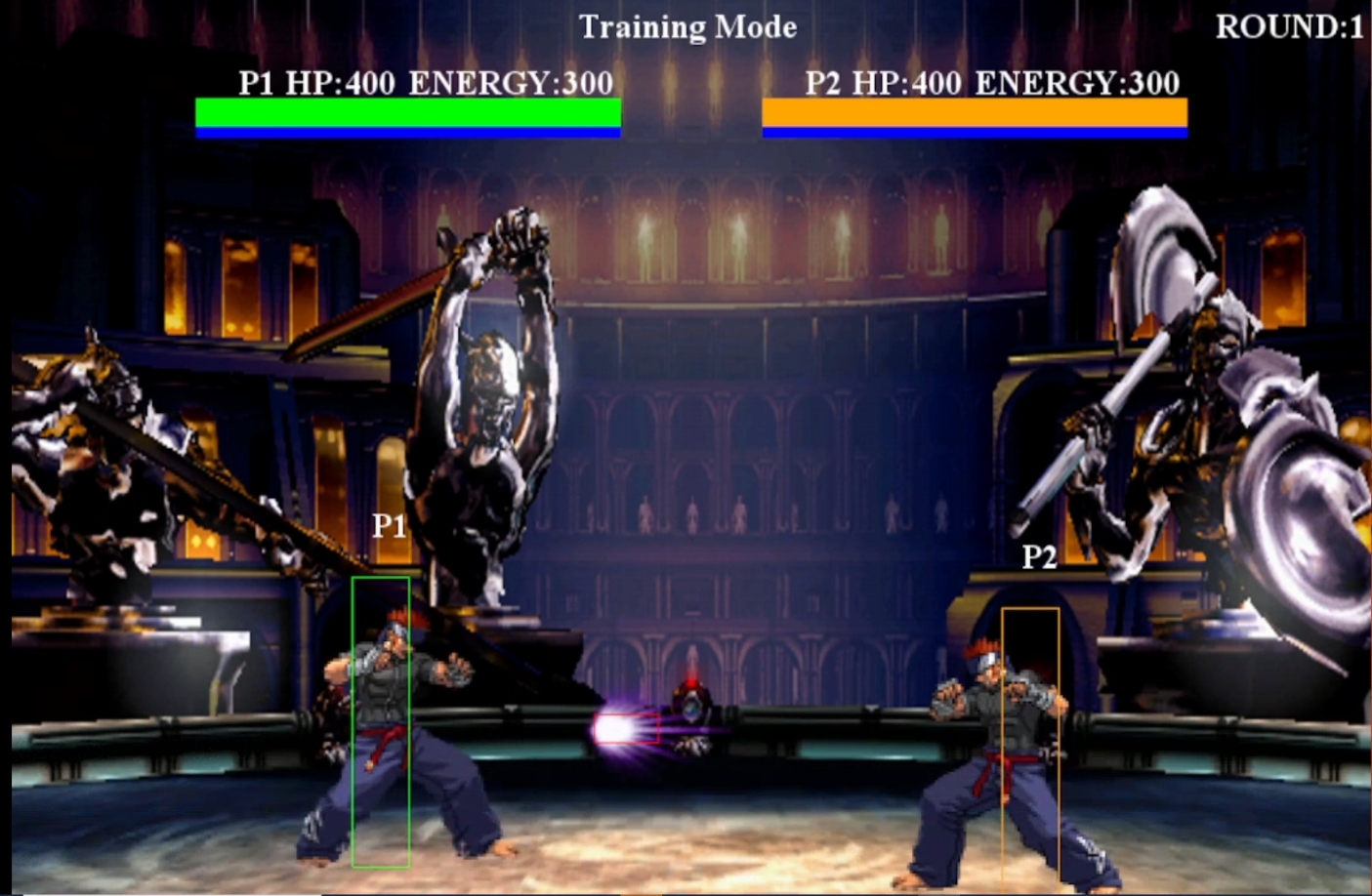}\hfill
    \includegraphics[width=.33\linewidth]{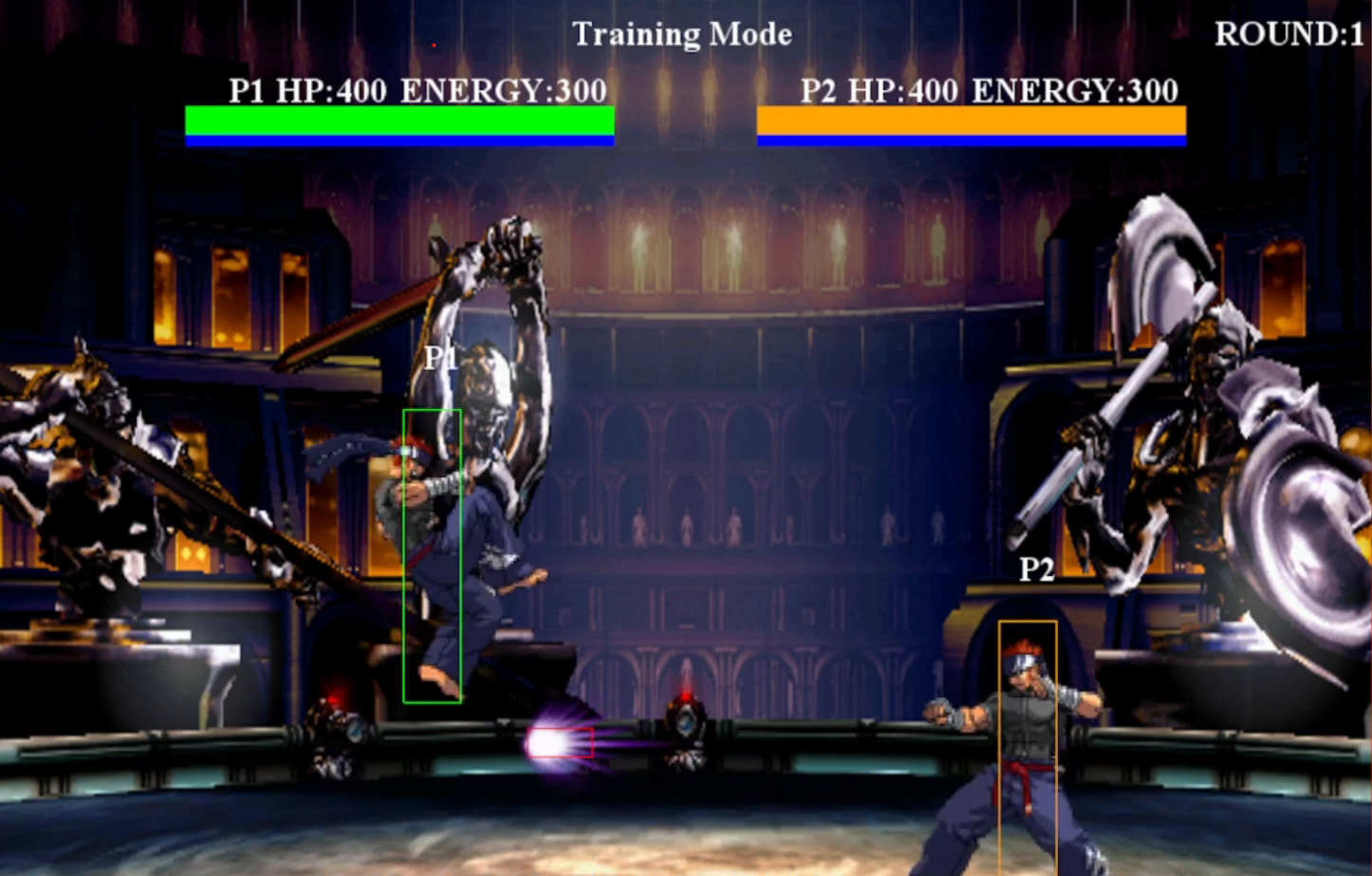}\hfill
    \includegraphics[width=.33\linewidth]{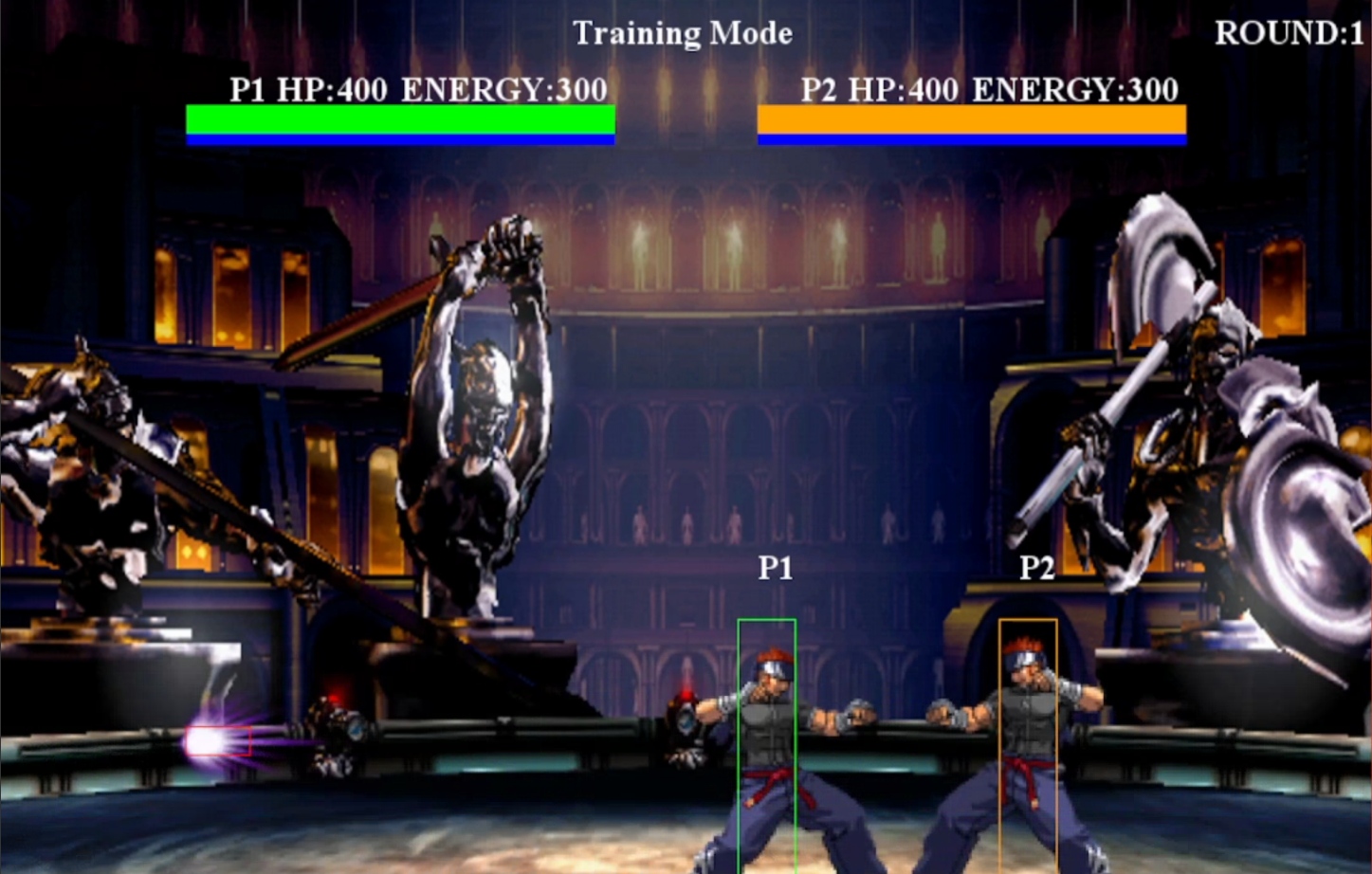}\hfill
    \caption{Our blind AI tries to avoid a fireball attack of the opponent AI in the DareFightingICE sound design.}
    \label{fig:fireballSequenceDareFightingiCE}
\end{figure*}
We conduct experiments on two different sound designs, the sound design of DareFightingICE, in the 2022 DareFightingICE Competition, and the sound design of FightingICE, in the 2021 FightingICE. The competition metrics mentioned in the previous section are used to evaluate these sound designs. Each combination of encoder and sound design is assessed with three trials of training and performance evaluation.

Table \ref{tblFightResults} shows the average fighting performance among three trials of our blind AI in each combination of sound design and audio encoder. The DareFightingICE sound design outperforms the FightingICE sound design on both performance metrics for each encoder. This was expected because the sound design of DareFightingICE is an enhanced version of its predecessor and targets visually impaired players, although there is room for improvement due to its nature of being a sample and baseline sound design for the 2022 competition.

Now, we discuss AI behaviors\footnote{Sample fight videos, where P1 is the blind AI and P2 is the opponent AI, of each of the six combinations of sound design and audio encoder are available on https://tinyurl.com/BlindAICoG2022.}. In particular, we focus on differences in the AI behaviors when the sound designs are those of DareFightingICE and FightingICE, both using the Mel-spectrogram encoder. The blind AI cannot avoid a fireball skill of its opponent in the FightingICE sound design because there are no sound cues when the skill is fired. On the contrary, because the sound cue is played when the opponent releases a fireball skill in the DareFightingICE sound design, the blind AI seems to be able to recognize the sound cue and tries to avoid the skill as much as possible. Figures \ref{fig:fireballFFTDareFightingICE} and  \ref{fig:fireballFFTFightingICE4.5} show the resulting Mel-spectrogram of audio data when the opponent AI attacks with a fireball action in the DareFightingICE sound design and the FightingICE sound design, respectively. The game screen sequence when our blind AI tries to avoid a fireball attack is shown in Fig. \ref{fig:fireballSequenceDareFightingiCE}.

The results above confirm that the proposed two metrics can be used to evaluate sound designs together with the evaluation done by the human judges in \cite{b6}. In the 2022 competition, the blind AI using the Mel-spectrogram encoder will be retrained from scratch to evaluate each entry sound design in the sound design track of the competition. In addition, the version trained in this work is made publicly available\footnote{https://tinyurl.com/DareFightingICE/SampleAI/BlindAI} as an official sample blind AI and will be used as a baseline AI in the AI track of the competition. 

\section{Conclusions}
In this paper, we introduced a blind AI that only uses sound as the input on the DareFightingICE platform. We also evaluated the performance of the AI with different audio encoders when it fought against an opponent AI whose performance had been tuned for visually impaired players in previous work. Our blind AI was able to beat the opponent AI when the FFT encoder or the Mel-spectrogram encoder was used. It was also found that the Mel-spectrogram encoder was the best.

For evaluation of sound designs in the DareFightingICE Competition, we proposed two metrics that are the win ratio and the average HP difference when fighting against the aforementioned opponent AI. Our experiment results showed that the sound design of DareFightingICE was more effective than the sound design of FightingICE. This confirms that the proposed two metrics can be used in evaluation of entry sound designs in the competition.

In the future, we plan to improve our blind AI to make it better understand the game state from sound observations, and to use the AI in research to procedurally generate effective sound designs.

\end{document}